# Crisis Domain Adaptation Using Sequence-to-sequence Transformers

**Congcong Wang**\*
School of Computer Science
University College Dublin
congcong.wang@ucdconnect.ie

**Paul Nulty**
School of Computer Science
University College Dublin
paul.nulty@ucd.ie

**David Lillis**
School of Computer Science
University College Dublin
david.lillis@ucd.ie

## ABSTRACT

User-generated content (UGC) on social media can act as a key source of information for emergency responders in crisis situations. However, due to the volume concerned, computational techniques are needed to effectively filter and prioritise this content as it arises during emerging events. In the literature, these techniques are trained using annotated content from previous crises. In this paper, we investigate how this prior knowledge can be best leveraged for new crises by examining the extent to which crisis events of a similar type are more suitable for adaptation to new events (cross-domain adaptation). Given the recent successes of transformers in various language processing tasks, we propose CAST: an approach for **C**risis domain **A**daptation leveraging **S**equence-to-sequence **T**ransformers. We evaluate CAST using two major crisis-related message classification datasets. Our experiments show that our CAST-based best run without using any target data achieves the state of the art performance in both in-domain and cross-domain contexts. Moreover, CAST is particularly effective in one-to-one cross-domain adaptation when trained with a larger language model. In many-to-one adaptation where multiple crises are jointly used as the source domain, CAST further improves its performance. In addition, we find that more similar events are more likely to bring better adaptation performance whereas fine-tuning using dissimilar events does not help for adaptation. To aid reproducibility, we open source our code to the community[1].

## Keywords

Domain Adaptation, Emergency Response, Social media, Transformers.

## INTRODUCTION

As evidenced by a number of previous research works (Imran, Mitra, et al. 2016; Alam et al. 2018; McCreadie et al. 2019), exploring computational techniques for finding useful information from user-generated content (UGC) on social media during crises remains an important research question. This is the case for two important reasons. Firstly, it is not easy to manually filter useful information since UGC is usually enormous and noisy (Imran, Castillo, et al. 2015), thus motivating the development of automatic filtering techniques. Secondly, since UGC contains a good deal of actionable information (e.g. a need for rescue following an earthquake), it has the potential to be utilised by emergency response agencies to aid these people in a timely manner (McCreadie et al. 2019).

Most approaches to crisis message classification require UGC from prior crises for training purposes (e.g. to fine-tune language models). The nature of such training data is the focus of this work. Crisis events of different types may feature similar characteristics. For example a bombing or an earthquake may both result in a person requiring rescue from a fallen building. Similarly, flooded conditions may arise from weather events such as hurricanes, or for other reasons such as a dam breach. As a new crisis unfolds, it is important to consider the training data that is used. Using training data from one crisis in order to classify posts from another is a form of cross-domain adaptation. Two primary research questions are to be addressed in this study. Firstly, to what extent does the similarity of crisis event types affect the quality of cross-domain adaptation in this context (one-to-one adaptation)? Secondly, although more training data is generally considered to be a positive, does this presumption hold when the training data relates to merging different crisis events with different characteristics (many-to-one adaptation)?

---
\*corresponding author
[1] https://github.com/wangcongcong123/CAST

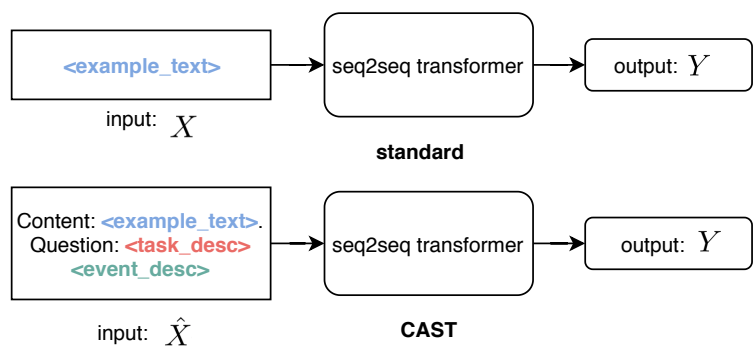

**Figure 1. The standard and CAST approaches for crisis domain adaptation leveraging seq2seq transformers**

In this work, we propose CAST that explores sequence-to-sequence (seq2seq) transformers for domain adaptation between different crises for message classification tasks. Unlike the standard method of fine-tuning seq2seq transformers for a downstream task (Raffel et al. 2020; Lewis et al. 2020), CAST is simple and trained by using a task description or/and event description added to each example, as illustrated in Figure 1. In comparison to similar work for crisis domain adaptation (H. Li, D. Caragea, et al. 2018; Alam et al. 2018; Liu et al. 2020), CAST uses only labeled source data without any unlabeled target data. This makes CAST more fit to real-world use cases. This is because a crisis usually focuses on a specific "topic" at a certain stage. For example, an earthquake is normally more about "Emerging Threats" than "Donation" at early stages. Hence, to get a good-quality distribution of the target event, the unlabeled target data needs to be collected as the target crisis unfolds.

To test the effectiveness of CAST, we conducted comparative experiments in both one-to-one (a single source event to a single target event) and many-to-one (several source events to a single target event) adaptation settings. In one-to-one adaptation, it is found that CAST outperforms the state of the art in both in-domain and cross-domain adaptation when not using any target data. To compare it with the standard, CAST is more effective than the standard in cross-domain adaptation when trained with a larger model. In the many-to-one adaptation, CAST's advantage over the standard in cross-domain adaptation becomes more obvious even with a small model. Moreover, based on our experimental figures, there is evidence to suggest that the use of multiple similar source events as the source domain improves the adaptation to a target event. Our main contributions are summarised as follows:

- We propose CAST: a seq2seq transformer-based approach for crisis domain adaptation. Our approach makes the model event-aware by taking into account a task descriptor and event descriptor in the input construction. In addition, it does not rely on any target data in model training which makes it more suitable for real-world use cases.

- In the one-to-one adaptation setting, experimental results show that CAST achieves competitive performance with existing work that uses unlabeled target data for self-supervision and it outperforms the state of the art when not using any target data. Moreover, we found CAST is particularly effective in cross-domain adaptation when trained with a larger model or combining multiple events as the source domain.

- Using CAST, we study the effectiveness of adaptation from similar events to a target event. The experimental results show that, to bring better adaptation performance, it is suggested to combine similar events as the source domain whereas adding dissimilar events does not help for adaptation.

## RELATED WORK

Our work aims to tackle the crisis domain adaptation problem by training computational models for crisis-related message classification. Hence, we surveyed related work from two perspectives: i), crisis domain adaptation and ii), crisis message classification.

**Crisis Domain Adaptation**

To overcome the issue of scarcity of training data for a new crisis, the problem of crisis domain adaptation has been widely studied in the literature (H. Li, Guevara, et al. 2015; Imran, Mitra, et al. 2016; H. Li, D. Caragea, et al. 2018; H. Li, X. Li, et al. 2018; Alam et al. 2018; X. Li and D. Caragea 2020). This line of work can be broadly divided into two categories: **target-data independent** and **target-data dependent**. The former is a supervised approach, where no unlabeled target data is used in model training. For example, Imran, Mitra, et al. (2016) investigated the domain adaptation between different combinations of past disasters across different languages. They found that similar events of the same type (e.g. earthquakes) tend to be useful for adaptation and even cross-language domain adaptation is useful when the source events and target events are from similar languages. Another target-data independent work is by H. Li, X. Li, et al. (2016), who explored a wide range of both word embeddings and sentence encodings with traditional machine learning (ML) algorithms for crisis adaptation classification tasks. They found that general pre-trained GloVe embeddings overall outperform other embeddings and the GloVe embeddings trained on crisis data bring better results on more specific crisis tweet classification tasks.

The other category is target-data dependent where classifiers are trained with labeled source data as well as unlabeled target data. Existing work in this category has consistently found that the adaptation performance can be improved by additionally using the unlabeled target data in model training (known as semi-supervised or self-training) (H. Li, Guevara, et al. 2015; H. Li, D. Caragea, et al. 2018; Alam et al. 2018; X. Li and D. Caragea 2020). For example, H. Li, D. Caragea, et al. (2018) trained Naïve Bayes classifiers for crisis domain adaptation with unlabeled target data taken into account via a self-training strategy. They compared the classifiers with their corresponding supervised classifiers learned only from labeled source data, showing the classifiers trained with extra unlabeled target data bring better adaptation performance. Moreover, they selected eleven event pairs for cross-domain adaptation study, presenting evidence that the adaptation between similar event pairs is likely to bring better performance. With the recent success of deep learning approaches based on neural networks (NNs) in short-message processing tasks (Y. Kim 2014), some work has been done to apply NN methods for the crisis adaptation problem. The representative work in this direction is from Alam et al. (2018). They applied a convolutional NN (CNN) with adversarial training and graph embeddings for domain adaptation between two crisis events in both supervised and semi-supervised (with unlabeled target data) settings, showing the semi-supervised way outperforms the supervised way. Another recent target-data dependent work is by X. Li and D. Caragea (2020). Instead of CNN, this work applied a recurrent neural networks (RNN) based seq2seq model that is trained jointly on a classification task with labeled source data and sequence reconstruction task with unlabeled target data. It is found that the reconstruction task can bring benefits to the adaptation performance as compared to the classification task alone.

Considering that even unlabeled target data is not readily available for a new crisis, CAST is a NN-based target-data independent approach, aiming to explore the limit of adaptation performance without knowing any target data.

**Crisis Message Classification**

To achieve the objective of finding useful information among UGC from social media, the literature has seen several works on classifying the messages by various "information types" (C. Caragea et al. 2011; Nguyen et al. 2017; Liu et al. 2020; Wang and Lillis 2020a). The information types exist in a wide range of forms. They can simply be binary indicating whether a message is relevant or informative to a certain disaster, or can be more fine-grained indicating different information nuggets such as requesting search and rescue, reporting infrastructural damage, etc. (Olteanu, Vieweg, et al. 2015; McCreadie et al. 2019). Given the importance of such classification tasks in emergency response, many computational techniques have been proposed for this purpose. The techniques vary from traditional ML algorithms to NN algorithms (C. Caragea et al. 2011; Nguyen et al. 2017). In particular, since the attention-based transformer NN architecture was introduced (Vaswani et al. 2017), recent years have witnessed great success of its variants (Devlin et al. 2019; Raffel et al. 2020), fine-tuned on downstream tasks. Two broad categories of the variants are encoder-based (e.g., BERT) and seq2seq-based (e.g., T5). Related work in both categories is found in the literature. For example, Liu et al. (2020) applied BERT for two crisis message classification tasks, leading to the state of the art performance. Per Wang and Lillis (2020), they leveraged the seq2seq T5 model for finding useful information from messages relating to the COVID-19 pandemic by treating a slot-filling classification task as a question-answering task. Our work is directly inspired by this work in constructing the input sequence with the addition of an event description. Since the core idea behind CAST is crisis domain adaptation, our work additionally takes into account the event type embedding (i.e, the event description) in the input construction. To the best of our knowledge, our work is the first to systematically study the problem of domain adaptation between different disasters by leveraging seq2seq transformer models for crisis message classification.

## METHOD

In this section, we first describe the background of fine-tuning seq2seq transformers for general downstream tasks and then introduce CAST for crisis domain adaptation based on this background.

### Seq2Seq

At a high level, a seq2seq model consists of a encoder and decoder. The encoder learns to encode an input example to a vector that can represent the contextualised linguistic features of the example. Conditional on the input representation, the decoder then learns to generate the prediction words iteratively. Mathematically, given a source sequence $X: \{x_1, x_2, ..., x_n\}$, the seq2seq model generates predictions denoted as the target sequence $\overline{Y}: \{\overline{y}_1, \overline{y}_2, ..., \overline{y}_m\}$ through a parameterised estimation of conditional probability distribution as follows.

$$p_{\theta_e, \theta_d}(\overline{\mathbf{Y}}_{1:m}|\mathbf{X}_{1:n}) = \prod_{i=1}^{m} p_{\theta_d}(\overline{\mathbf{y}}_i|\overline{\mathbf{Y}}_{0:i-1}, f_{\theta_e}(\mathbf{X}_{1:n})) \quad (1)$$

Where $f_{\theta_e}(\cdot)$ refers to the mapping function from the source sequence $X$ to its contextualised representation, learnt by the encoder with tunable parameters $\theta_e$. Likewise, $\theta_d$ is the tunable parameters for the decoder to learn the function of conditional generation: $p_{\theta_d}(\cdot)$. In order to optimise $\theta_e$ and $\theta_d$, the model is trained with the objective function defined as follows.

$$\arg\min_{\theta_e, \theta_d} \sum_{i=1}^{m} \xi_{\text{cross\_entropy}}(y_i, \overline{y}_i) \quad (2)$$

As can be seen, $\theta_e$ and $\theta_d$ are tuned with the objective of minimising the cross entropy loss between the ground truth targets $Y: \{y_1, y_2, ..., y_m\}$ and the predicted targets $\overline{Y}: \{\overline{y}_1, \overline{y}_2, ..., \overline{y}_m\}$. In the fine-tuning of a downstream classification task, $\theta_e$ and $\theta_d$ are first initialised from their corresponding pre-trained parameters and then tuned on the objective function where the $Y: \{y_1, y_2, ..., y_m\}$ refers to the ground truth labels of the task.

### CAST

The problem of domain adaptation between crisis events can be simply described as follows. Given a task $\mathcal{T}$, for a set of source events $S$ with dataset $S_d$, a model trained on $S_d$ is directly tested on the test set $T_d$ of a set of target events $T$ within the same task $\mathcal{T}$. Referring to the aforementioned seq2seq model as the model mentioned here, this means that $\theta_e$ and $\theta_d$ are learned to fit the source dataset $S_d$ through fine-tuning the model first and then the model does inference directly on the target dataset $T_d$ without further training. To differentiate **in-domain** and **cross-domain** adaptation, the former refers to the case when $S$ equals $T$ while the latter is the case when $S$ is not intersected with $T$, which is the focus of this study. To put the cross-domain adaptation in the context of crisis response, the source events set $S$ refers to the past crises whose datasets are available and the target events set $T$ usually contains one element referring to an emerging new crisis.

For the standard method of fine-tuning a seq2seq model for a downstream task as seen in (Raffel et al. 2020; Lewis et al. 2020), the input example $X: \{x_1, x_2, ..., x_n\}$ simply consists of the textual content itself (see Figure 1). CAST is specifically proposed for cross-domain adaptation, taking into account both a task description $T_{desc}$ and an event description $E_{desc}$, leading to the new input $\hat{X}: \{\hat{x}_1, \hat{x}_2, ..., \hat{x}_k\}$, formulated as follows.

$$\hat{X}_{1:k} = \zeta_\oplus(X_{1:n}, T_{desc}, E_{desc}) \quad (3)$$

Where $\zeta_\oplus$ is the transformation function that concatenates $X$ with $T_{desc}$ and $E_{desc}$ in a natural language form. For example, in Figure 1, $\hat{X}$ is constructed as a question-answering sequence. Following this, Equation 1 now becomes:

$$p_{\theta_e, \theta_d}(\overline{\mathbf{Y}}_{1:m}|\hat{\mathbf{X}}_{1:k}) = \prod_{i=1}^{m} p_{\theta_d}(\overline{\mathbf{y}}_i|\overline{\mathbf{Y}}_{0:i-1}, f_{\theta_e}(\hat{\mathbf{X}}_{1:k})) \quad (4)$$

As described, CAST differs from the standard approach in two main aspects. First, it considers $T_{desc}$, which is inspired by prior work (Wang and Lillis 2020b) using a task description in the input example for a COVID-related event extraction task. In addition, CAST considers $E_{desc}$ for making the model domain-aware when tested on different events.

## EXPERIMENTS

In this section, we describe experimental details, report and discuss the results from different dimensions, aiming to comprehensively test the effectiveness of CAST and share the insights from what we have learned.

## Data Preparation

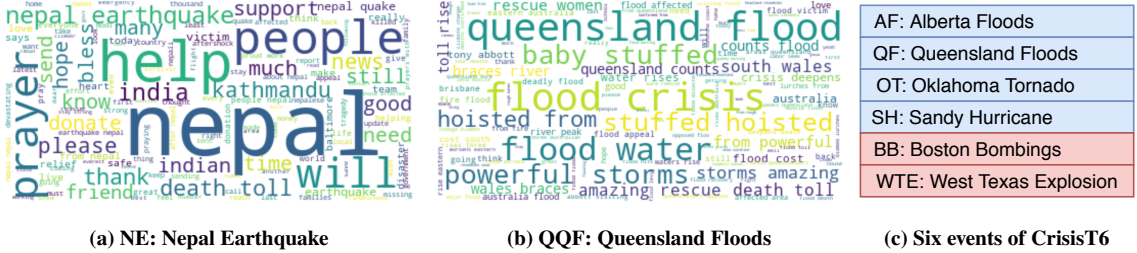

(a) NE: Nepal Earthquake     (b) QQF: Queensland Floods     (c) Six events of CrisisT6

**Figure 2.** Subfigure (a) and (b) show the word clouds of examples of the two events from the nepal_queensland dataset and (c) presents the six crisis events of CrisisT6.

Since CAST is proposed for cross-domain adaptation between crises, we use datasets containing examples from different crisis events. The datasets we used in our experiments are **nepal_queensland** and **CrisisT6**, described briefly as follows.

- **nepal_queensland** is a famous benchmark dataset for cross-domain adaptation in this field, consisting of two crisis events, Nepal Earthquake and Queensland Floods whose word distributions are depicted in the word clouds of Figure 2a and 2b. The samples of this dataset are tweets related to the two events and each tweet is annotated by two well-balanced classes: *relevant* implying the tweet is relevant to the event and *not_relevant* implying the opposite. We use the standard train, validation and test splits from Alam et al. (2018) in our experiment for parallel comparison.

- **CrisisT6** was originally released by Olteanu, Castillo, et al. (2014). It is a collection of approximately 60,000 tweets posted during six crisis events with approximately 10,000 in each event. Six events are presented in Figure 2c. Each tweet in each event of this dataset is labeled with two classes, *on-topic* indicating the tweet is on-topic to the event and *off-topic* indicating the opposite and the two classes are well-balanced across the examples. Since CrisisT6 does not provide standard splits, following H. Li, D. Caragea, et al. (2018), we use 5-fold cross-validation evaluation whenever the in-domain performance is reported for CrisisT6.

Considering both datasets relate to a similar task, (i.e., $\mathcal{T}$ is defined as a binary classification task), we unify their labels to the same target labels. In **nepal_queensland**, *relevant* is changed to *yes* and *not_relevant* is changed to *no*, likewise for **CrisisT6**. Following this unification, the task description $T_{desc}$ becomes the same for the two datasets.

## Scenarios

Using the datasets, our experiments include two scenarios for training, as outlined in Figure 1.

- **standard**: This scenario represents common practice in the literature for fine-tuning seq2seq transformers on a downstream task. It is used as a strong baseline in our experiments due to its state of the art performance on text classification tasks (Raffel et al. 2020). This scenario simply feeds the raw training examples to the model without any additional text being added. In our case, a training example no matter what crisis event it belongs to is always constructed in the form: " {*tweet_text*}".

- **postQ**: This scenario represents a particular use case of our CAST method. Referring back to Equation 3, postQ takes into account $T_{desc}$, $E_{desc}$ and $\zeta_\oplus$. In this scenario, $T_{desc}$ becomes "Is this message relevant to" and $E_{desc}$ becomes {*location_name*}{*crisis_name*} that are available in our selected datasets [2]. Finally, the concatenation function $\zeta_\oplus$ constructs the input example to be in the form of a question-answering sequence: "Content: {*tweet_text*}. Question: Is this message relevant to {*location_name*}{*crisis_name*}?". For in-domain adaptation, the location name and crisis name are from the source event(s) at both training and testing time. For cross-domain adaptation, the location name and crisis name are from the source event(s) at training time and from the target event at testing time.

---

[2] For example, in the nepal_queensland dataset, $E_{desc}$ becomes "Nepal Earthquake" or "Queensland Floods".

Before determining postQ, we also experimented with different variants of CAST with CrisisT6. These are summarised as follows:

- **variant 1**. This is similar to postQ except that we remove the {*location_name*} from $E_{desc}$, making the input location-agnostic to the model. The final input sequence is constructed like: "Content: {*tweet_text*}. Question: Is this message relevant to {*crisis_name*}?".

- **variant 2**. This is similar to postQ except that we re-arrange {*location_name*} and {*crisis_name*} such that the input is like: "Content: {*tweet_text*}. Question: Is this message relevant to a {*crisis_name*} event that occurred in {*location_name*}?".

- **variant 3**. This variant constructs the input sequence by setting $T_{desc}$ to be empty such that the input is like: "Content: {*tweet_text*}. Question: {*location_name*}{*crisis_name*}?".

The experimentation on these variants and postQ did not present any noticeable difference in performance. It is interesting to notice that there is no performance difference between variant 3 and postQ where variant 3 simply uses location and crisis name without including $T_{desc}$ in the extended text. This is because given a specific classification task, $T_{desc}$ will be the same for all training examples, thus leading to no difference to the model. However, we ultimately chose the variant with $T_{desc}$ (i.e., postQ) in our subsequent experiments mainly because we expect to expand our approach to multi-task learning settings as future work where $T_{desc}$ becomes different for different tasks.

**Training**

Our experiments are conducted to examine the performance of the above two scenarios in domain adaptation through fine-tuning seq2seq transformers on the two benchmark datasets. Given a number of existing such seq2seq models (Wolf et al. 2020), we select T5 (Raffel et al. 2020) as the target model in our study due to the availability of multiple pre-trained weights and its strong performance in various downstream language tasks. To be specific, the off-the-shelf `t5-small` and `t5-base` weights implying different model sizes are used in our study, which we abbreviate to `small` and `base`[3]. In fine-tuning, we configure most of the hyper-parameters in keeping with related work (Wang and Lillis 2020b). We fine-tuned both the `small` and `base` models with 12 training epochs as we see no further improvement when training with more epochs [4]. The learning rate is set to be $5e − 05$ using Adam optimizer (Kingma and Ba 2015), updated by a linear decay scheduler with warmup ratio 10% of the total training steps. All our experimental runs are accelerated by a 6GB RTX2060 GPU, so we adopt memory saving strategies including Mixed Precision Training (FP32 and FP16) (Micikevicius et al. 2018) and gradient accumulation steps (up to 4) to ensure the training batch size is always 16. In addition, we set the maximum source and target sequence length to be 128 and 10 since we are in the context of processing short crisis messages that do not exceed this length.

**Results and Discussions**

Having conducted extensive experiments with the two selected benchmark datasets, we report the results regarding both in-domain and cross-domain adaptation. To align with the metrics used in the literature, the weighted F1 scores are reported for **nepal_queensland** (Alam et al. 2018) and the accuracy scores are reported for **CrisisT6** (Liu et al. 2020). We report and discuss the experimental results from the following two perspectives, which is inspired by the intuition behind real-world crisis domain adaptation.

- **One-to-one adaptation** is when both the source events set *S* and the target events set *T* relate to a single one crisis event. This helps answer a question like: "Among a number of source events whose training sets are available, which one is most suitable to be adapted to a new emerging target event for which training data is not yet available?".

- **Many-to-one adaptation** is similar to the one-to-one adaptation except that the source events set *S* can contain multiple events. It helps answer a question like: "Which combination of available source events, is most suited to be adapted to an emerging target event?".

|                        | NE→NE | QQF→QQF |
|------------------------|-------|---------|
| CNN (Alam et al. 2018) | 65.11 | 93.54   |
| **standard-small**     | 75.35 | 96.31   |
| **standard-base**      | 74.40 | 96.83   |
| **postQ-small**        | 79.25 | 96.34   |
| **postQ-base**         | 77.57 | 96.81   |

**Table 1.** The in-domain adaptation weighted F1 scores for **nepal_queensland** where NE: Nepal Earthquake and QQF: Queensland Floods. The CNN runs refers to the supervised CNN run from Alam et al. (2018) and the runs in bold refer to our experimental standard and postQ runs with `t5-small` and `t5-base`.

|     | source → target                     | NE→QQF | QQF→NE |
|-----|-------------------------------------|--------|--------|
| TDD | CNN + DA + GE (Alam et al. 2018)    | 65.92  | 59.05  |
|     | RNN + AE (X. Li and D. Caragea 2020)| 81.18  | 68.38  |
| TDI | CNN + DA (Alam et al. 2018)         | 60.94  | 57.79  |
|     | RNN (X. Li and D. Caragea 2020)     | 55.17  | 64.18  |
|     | **standard-small**                  | 82.43  | 58.99  |
|     | **standard-base**                   | 77.39  | 60.25  |
|     | **postQ-small**                     | 78.21  | 63.75  |
|     | **postQ-base**                      | 87.06  | 64.12  |

**Table 2.** The cross-domain adaptation weighted F1 scores for **nepal_queensland**. The runs are presented in two categories: target-data-dependent (TDD) and target-data-independent (TDI). The CNN+DA+GE refers to the CNN run with data adversarial with graph embedding from Alam et al. (2018) and RNN+AE refers to the RNN auto-encoder run from X. Li and D. Caragea (2020).

*One-to-one Adaptation*

Table 1 and 2 present the in-domain and cross-domain performance respectively on the **nepal_queensland** dataset across different runs. Regarding the in-domain performance, we compare our runs with the CNN run (Alam et al. 2018). It is found that our runs substantially outperform this run in both NE and QQF events. For cross-domain adaptation, we include CNN+DA+GE (Alam et al. 2018) and RNN+AE (X. Li and D. Caragea 2020) that use unlabeled target data in training, i.e., target-data-dependent (TDD). Apart from our target-data independent (TDI) runs, we also report CNN+DA and RNN that do not use any target data like our runs. To compare our runs with the TDD runs, our runs substantially outperform the CNN+DA+GE run. When it comes to the state of the art (SOTA) TDD RNN+AE run, our postQ-base achieves competitive performance, where 87.06 versus 81.18 in NE→QQF and 64.12 versus 68.38 in QQF→NE. However, the postQ-base outperforms the SOTA TDI RNN run in cross-domain adaptation where 87.06 versus 55.17 in NE→QQF and 64.12 versus 64.18 in QQF→NE. To compare the standard with the postQ in in-domain adaptation (see Table 1), we found that the postQ performs basically the same as the standard. This makes sense since postQ is only different from standard in appending an extra task and event description. For in-domain adaptation, the appended text is the same for all training examples during training and at inference time, thus leading to no difference for the model when the extra text exists or not, which explains why they have the same level of performance. Interestingly for cross-domain adaptation, the standard method can achieve comparable performance to our postQ when using a small model. When using a bigger model (i.e., postQ-base), it seems that the standard does not maintain comparable performance and instead our postQ performs much better (see Table 2). This finding is further verified in our subsequent experiments on the **CrisisT6** dataset.

Based on the **nepal_queensland** dataset, we have identified some evidence of the effectiveness of our CAST-based runs (particularly for TDI cross-domain adaptation) as compared to the SOTA. However, one limitation of this benchmark dataset is that it only offers two crisis events, which limits its value in fully evaluating the effectiveness of our approach. Hence, we extend our experiments to be conducted upon the **CrisisT6** dataset that consists of six different crisis events representing a wide range of domains.

---

[3]`t5-small` and `t5-base` have around 60M and 220M parameters respectively. There are larger versions originally released by the authors, such as `t5-large`, `t5-3B` and `t5-11B`. We did not include these since they are too large to be handled by our training resources.

[4]Except that we fine-tuned `base` on **nepal_queensland** with 6 epochs.

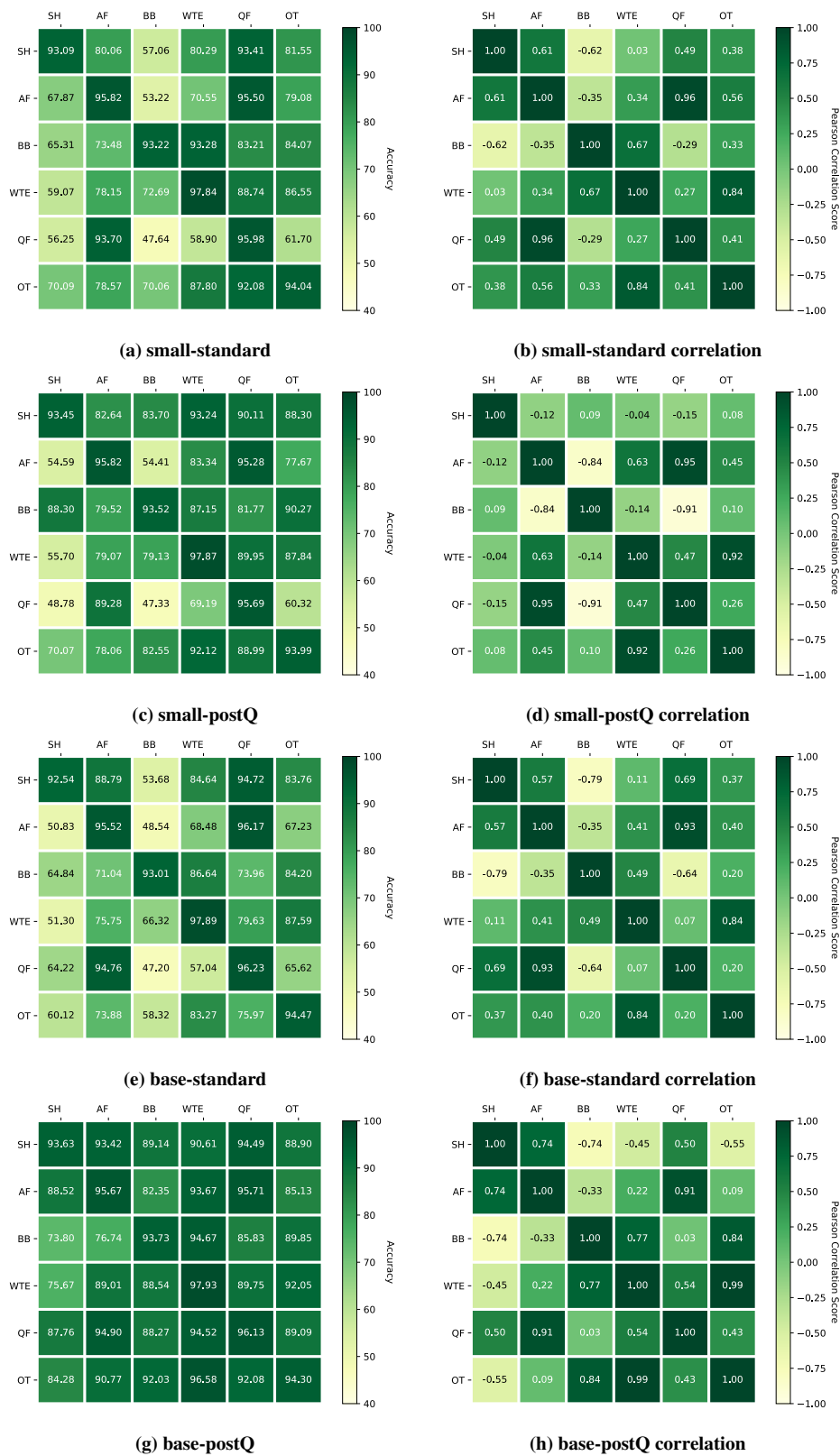

**Figure 3.** Event adaptation accuracy and correlation for CrisisT6 where SH: Sandy Hurricane, AF: Alberta Floods, BB: Boston Bombing, WTE: West Texas Explosion, QF: Queensland Floods, and OT: Oklahoma Tornado. Figure (a), (c), (e) and (g) refer to the adaptation accuracy from the source events (rows) to the target events (columns) using the standard and postQ two ways of fine-tuning `small` and `base`. Subfigure (b), (d), (g), and (h) demonstrate the Pearson Correlation matrix between the rows of subfigure (a), (c), (e) and (g) respectively.

|  | SH→QF | SH→BB | SH→WTE | SH→OT | SH→AF | QF→BB |
|---|---|---|---|---|---|---|
| NB-S | 76.84 | 68.66 | 77.21 | 80.78 | 71.06 | 74.97 |
| NB-ST | 82.40 | 84.06 | 90.82 | 87.76 | 82.57 | 81.86 |
| **postQ-base** | 94.49 | 89.14 | 90.61 | 94.49 | 93.42 | 88.27 |

|  | QF→OT | QF→AF | BB→OT | BB→AF | BB→WTE | Average |
|---|---|---|---|---|---|---|
| NB-S | 84.13 | 84.35 | 73.81 | 78.87 | 94.77 | 78.68 |
| NB-ST | 85.48 | 86.91 | 83.96 | 86.01 | 94.82 | 86.06 |
| **postQ-base** | 89.09 | 94.90 | 89.85 | 76.74 | 94.67 | 90.52 |

**Table 3.** Cross-domain accuracy comparison between our postQ-base, NB-S (TDI) and NB-ST (TDD) from H. Li, D. Caragea, et al. (2018) who studied the 11 event pairs of CrisisT6 as reported in the table.

Figure 3 reports the results of domain adaptation between the six events using both the standard method and postQ for fine-tuning the small and base models. The subfigures on the left side, i.e., (a), (c), (e), (g), present the accuracy scores in a matrix where the diagonals refer to the in-domain performance and the rest are cross-domain scores. The row represents the source events while the column stands for the target events. In addition to the adaptation matrices, we also add a correlation matrix to the right side of each of the adaptation matrices. The correlation matrices calculate the Pearson correlation between the rows of source events, which helps indicate how correlated two source events are in terms of their applicability to other target events.

Examining the matrices on the left, we find that both the in-domain and cross-domain performance is consistent with **nepal_queensland**. For example, in in-domain adaptation, postQ achieves similar performance to the standard with different model sizes[5]. Speaking of cross-domain adaptation, postQ slightly outperforms the standard when using the `small` model (see Figure 3a and 3c). For the `base` model, the postQ substantially outperform the standard in cross-domain adaptation (see Figure 3e and 3g). Since H. Li, D. Caragea, et al. (2018) has done a similar work on this dataset, we compare our postQ-base with their NN-S and NB-ST runs on 11 event pairs, as presented in Table 3. It shows that the postQ-base substantially outperforms the target-data independent NB-S which is consistent to the results we report for **nepal_queensland**. However, the postQ-base achieves strong performance overall across the 11 pairs adaptation as compared to the target-data dependent NB-ST (90.02 versus 86.06 in average accuracy).

Regarding cross-domain adaptation, we also noticed some interesting points. The results from H. Li, D. Caragea, et al. (2018) present some evidence that similar event pairs such as QF→AF and BB→WTE are more likely to bring better scores than dissimilar pairs like QF→BB and BB→AF (see Table 3). This evidence is enhanced in our study. We noted that the Alberta Floods (AF) and Queensland Floods (QF) relate to the same type of crisis (flooding) albeit in different locations at different times. It is interesting that in our four runs these two events are **reciprocal**, indicating that either of them as the source event is well-suited to being adapted for the other as the target event. For example, the AF→QF adaptation always achieves accuracy around 95 and QF→AF adaptation achieves accuracy of 89.28 at the worst (Figure 3c). Examining their correlation scores on the right, we find that they are not only reciprocal, but also **highly correlated**, ranging from 0.91 to 0.96 (Figure 3h and 3b). This lends credence to the idea that similar event types have similar characteristics in terms of their applicability to cross-domain adaptation and could potentially be used interchangably for a novel target event.

Another event pair with some similar characteristics are the Boston Bombing (BB) and the West Texas Explosion (WTE). These are perhaps less similar to the floods above in that one was an intentionally planted explosive device whereas the other was a factory fire that later resulted in an explosion. In this situation, it can be seen that whereas BB can be successfully adapted to WTE, the reverse is not the case. This may be related to the observation that BB is itself a difficult target event to adapt to, as evidenced by the fact that the cross-domain adaptation tends to be poorest in general when BB is the target.

Surprisingly, we find that the Sandy Hurricane (SH) and Oklahoma Tornado (OT) datasets can not only be well adapted to AF and QF in most cases but also adapt well to WTE. This implies that there may be a certain degree of common linguistic features shared between the tornado/hurricane events and the explosion event. It seems that these findings indicate that the more similar a source event is to a target event, the more likely it is to exhibit better adaptation performance for that target event. Now such a question is naturally raised: does combining multiple similar source events add further benefit (many-to-one adaptation)?

---

[5]Per Liu et al. (2020) that reported the average in-domain SOTA accuracy for CrisisT6 which is 95.6, our in-domain scores as seen from the diagonals approximate this SOTA. Since this SOTA was gained based on a random test leaving-out evaluation which is different from our 5-fold validation, we only refer it here instead of comparing directly it with our scores.

| | small-standard | | small-postQ | |
|---|---|---|---|---|
| AF+BB+WTE+QF+OT→SH | 74.35 | | AF+BB+WTE+QF+OT→SH | 82.16 |
| SH+BB+WTE+QF+OT→AF | 95.22 | | SH+BB+WTE+QF+OT→AF | 95.16 |
| SH+AF+WTE+QF+OT→BB | 70.18 | | SH+AF+WTE+QF+OT→BB | 88.01 |
| SH+AF+BB+QF+OT→WTE | 89.39 | | SH+AF+BB+QF+OT→WTE | 95.94 |
| SH+AF+BB+WTE+OT→QF | 95.59 | | SH+AF+BB+WTE+OT→QF | 95.71 |
| SH+AF+BB+WTE+QF→OT | 90.52 | | SH+AF+BB+WTE+QF→OT | 89.16 |
| Average | 85.88 | | Average | 91.03 |

**Table 4.** Leave-one-out cross-domain adaptation accuracy using CrisisT6. The last row reports the average score. As a comparison, the best average score reported in the literature is 89.6 (H. Li, X. Li, et al. 2018).

*Many-to-one Adaptation*

Considering the variety of crisis events and training efficiency, our many-to-one experimental runs are based on the **CrisisT6** dataset trained with the `small` model. The first experiment we conduct is leave-one-out cross-domain adaptation where we choose only one crisis as the target domain and the union of the others as the source domain. Table 4 presents the results of fine-tuning using both the standard and postQ approaches. To compare our runs with the work by H. Li, X. Li, et al. (2018), it reveals that our CAST-based postQ run achieves 91.03 versus 89.6 in average accuracy. In addition, from this table it can be seen that there is no substantial difference between standard and postQ when AF, QF and OT are left out (they already achieve high accuracy). However, when leaving out SH and BB, standard performance is substantially lower than for postQ. This experiment indicates, even with the `small` model, that postQ outperforms the standard approach when considering multiple events as the source domain. This is further justified by our next experiment.

Our next experiment is conducted to test what effect of enriching the source domain by combining multiple events. For this purpose, we select two event pairs: (QF, AF) and (BB, WTE). As indicated above, each pair contains two similar event types, while the two pairs themselves are dissimilar. The decision to choose these two pairs as similar event pairs is guided by existing work (H. Li, D. Caragea, et al. 2018) and the correlation scores that are discussed above.

Table 5 presents the results of combinations of multiple source evens adapted to the two pairs. First, we see from the figures that postQ overall outperforms the standard in most situations (accuracy is much higher when BB and WTE are the target events and is at least the same level for QF and AF). However, the more interesting thing we note from this experiment is that simply increasing the number of source events does not guarantee benefits to the adaptation performance but it depends on what source events to be added. For example, when QF is the target event, we see only a trivial difference between AF-to-QF and leaving QF out (thus combining all other crises as the source domain). A similar pattern is observed when AF is the target event.

| | small-standard | | small-postQ | |
|---|---|---|---|---|
| | QF | AF | QF | AF |
| AF | 95.5 | - | 95.28 | - |
| QF | - | 93.7 | - | 89.28 |
| BB+WTE | 83.79 | 74.69 | 87.27 | 76.18 |
| SH+OT | 93.97 | 85.05 | 93.71 | 83.71 |
| AF+SH+OT | 95.32 | - | 95.76 | - |
| QF+SH+OT | - | 95.45 | - | 95.38 |
| SH+OT+BB+WTE | 92.72 | 82.67 | 91.44 | 81.81 |
| AF+SH+OT+BB+WTE | 95.59 | - | 95.71 | - |
| QF+SH+OT+BB+WTE | - | 95.22 | - | 95.16 |

(a) QF and AF as the target events

| | small-standard | | small-postQ | |
|---|---|---|---|---|
| | BB | WTE | BB | WTE |
| WTE | 72.69 | - | 79.13 | - |
| BB | - | 93.28 | - | 87.15 |
| AF+QF | 51.24 | 68.42 | 50.1 | 80.56 |
| SH+OT | 75.71 | 89.32 | 82.14 | 92.01 |
| WTE+SH+OT | 73.3 | - | 85.3 | - |
| BB+SH+OT | - | 92.44 | - | 96.35 |
| AF+QF+SH+OT | 63.94 | 79.6 | 78.42 | 86.29 |
| WTE+AF+QF+SH+OT | 70.18 | - | 88.01 | - |
| BB+AF+QF+SH+OT | - | 89.39 | - | 95.94 |

(b) BB and WTE as the target events

**Table 5.** Many-to-one cross-domain adaptation on similar event pairs

Table 5a also indicates that adding BB+WTE seems not to add any benefit to the performance (indeed this reduces performance when compared with a source domain of SH+OT), and SH+OT can help to a degree (adding SH+OT to QF results in an improved cross-domain performance when AF is the target).

Table 5b demonstrates a similar outcome. When BB and WTE are the target events, AF+QF contributes little to the adaptation performance. Surprisingly, SH+OT not only helps QF and AF but also helps BB and WTE, which coincides with the one-to-one results as reported in the previous section. Hence, as a recommendation to maximise the adaptation performance to a target event (e.g., AF), it is good to combine its similar events (e.g., QF+SH+OT → AF) as the source domain and exclude dissimilar events (e.g., BB+WTE) [6].

## CONCLUSION AND FUTURE WORK

In this work, we propose CAST - a sequence-to-sequence (seq2seq) transformer-based approach for domain adaptation between crisis events. To test the effectiveness of CAST, we conduct extensive experiments on two benchmark crisis-related messages classification datasets. In one-to-one adaptation setting, CAST is demonstrated to be effective in cross-domain adaptation outperforming the state of the art without using any target data and its advantage over the standard approach is more pronounced when training with a bigger model. In a many-to-one adaptation setting with a small model, as compared to the standard method, CAST adds substantial improvements to the cross-domain adaptation performance.

Interestingly, our results indicate that for cross-domain adaptation there is merit in choosing a source domain with similar characteristics (i.e. fine-tuning based on a similar type of crisis). If multiple existing similar events are available, these can be combined to form a larger source dataset to improve adaptation performance. Dissimilar events may harm classification performance, however. Regarding future work, so far our method has only been tested on the binary relevance classification tasks for crisis messages. We intend to test our method in a wider range of tasks such as eyewitness identification (Zahra et al. 2020) and more fine-grained information type classification (Olteanu, Vieweg, et al. 2015; McCreadie et al. 2019). In addition, as our method includes a task description for model training, we also intend to extend our approach to domain adaptation in a multi-task learning setting, extending our work in Wang and Lillis (2021).

## ACKNOWLEDGMENTS

This work was supported by the Enterprise Ireland and European Union Career-FIT programme under the Marie Sklodowska-Curie grant agreement No. 713654

---

[6] Although there is no direct harm to the performance when including the dissimilar events (see the last two rows of Table 5a and 5b), it is suggested to exclude them which can help reduce the training size and thus improve the training efficiency.